\documentclass[11pt]{article}

\usepackage[preprint]{acl}

\usepackage{times}
\usepackage{latexsym}
\usepackage{graphicx}

\usepackage[T1]{fontenc}
\usepackage[utf8]{inputenc}

\usepackage{microtype}

\usepackage{inconsolata}

\usepackage{graphicx}
\usepackage{multirow}
\usepackage{booktabs}
\usepackage[utf8]{inputenc}
\usepackage{CJKutf8}

\usepackage{hyperref}
\title{Causal Tree Extraction from Medical Case Reports:\\A Novel Task for Experts-like Text Comprehension}

\author{
  Sakiko Yahata\textsuperscript{1} Zhen Wan\textsuperscript{1} Fei Cheng\textsuperscript{1} Sadao Kurohashi\textsuperscript{1} Hisahiko Sato\textsuperscript{2} Ryozo Nagai\textsuperscript{3} \\
  \textsuperscript{1}Kyoto University, \textsuperscript{2}Precision Inc., \textsuperscript{3}Jichi Medical University \\
  \textsuperscript{1}\texttt{\{yahata, ZhenWan, feicheng, kuro\}@nlp.ist.i.kyoto-u.ac.jp} \\\textsuperscript{2}\texttt{satoh@premedi.co.jp} \textsuperscript{3}\texttt{rnagai@jichi.ac.jp} \\
}

\begin{document}
\maketitle
\begin{abstract}
Extracting causal relationships from a medical case report is essential for comprehending the case, particularly its diagnostic process.
Since the diagnostic process is regarded as a bottom-up inference, causal relationships in cases naturally form a multi-layered tree structure.
% This leads to existing tasks, such as medical relation extraction, are insufficient for capturing the causal relationships of an entire case, as they treat each relation triplet equally without considering the hierarchical structure inherent in the diagnostic process.
The existing tasks, such as medical relation extraction, are insufficient for capturing the causal relationships of an entire case, as they treat all relations equally without considering the hierarchical structure inherent in the diagnostic process.
Thus, we propose a novel task, Causal Tree Extraction (CTE), which receives a case report and generates a causal tree with the primary disease as the root, providing an intuitive understanding of a case's diagnostic process.
% A causal tree provides us an intuitive understanding of the entire mechanism of a case and highlighting important parts, not only insight into a diagnostic reasoning procedure.
% A causal tree provides us an intuitive understanding of the entire mechanism of a case, particularly the important parts in the diagnostic process.
Subsequently, we construct a Japanese case report CTE dataset, J-Casemap, propose a generation-based CTE method that outperforms the baseline by 20.2 points in the human evaluation, and introduce evaluation metrics that reflect clinician preferences.
Further experiments also show that J-Casemap enhances the performance of solving other medical tasks, such as question answering.
% }

\end{abstract}

\section{Introduction}
% 症例報告の因果木要約
% 専門家のような文章理解のための要約タスク

% \begin{figure}[t!]
%     \centering
%     \includegraphics[width=0.92\linewidth]{TSS.eps}
%     \caption{Medical causal tree extraction. The layer number corresponds to the depth of the node. The English medical terms are translated with GPT-4o for demonstration only.}
%     \label{fig:CTE}
% \end{figure}
\begin{figure}[t!]
    \centering
    \includegraphics[width=\linewidth]{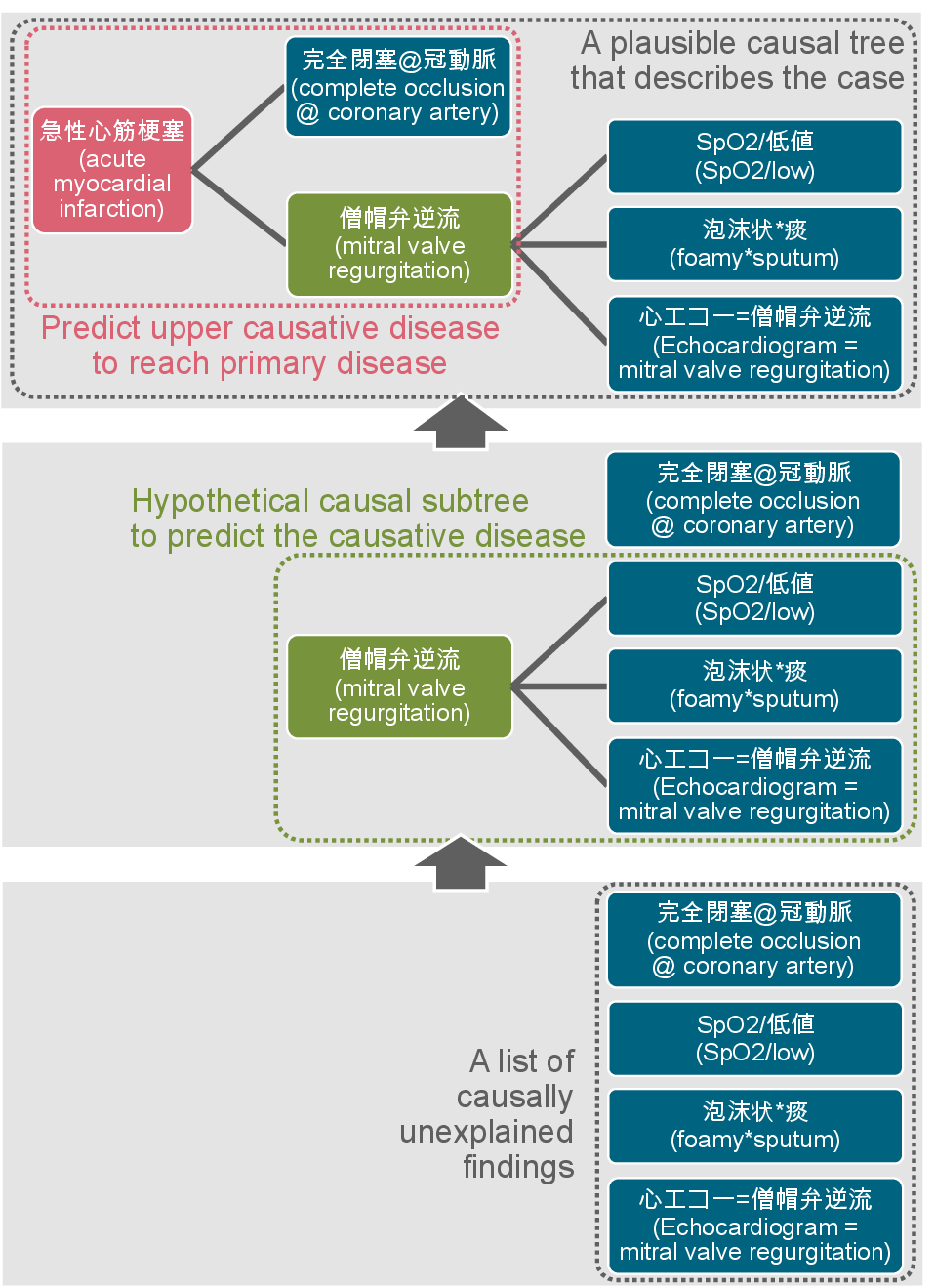}
    \caption{A diagnostic bottom-up procedure.}
    \label{fig:diagnosis-flow}
\end{figure}

% 最初にテキスト理解
% 次にテキスト理解のための因果の重要性

% \textcolor{blue}{
% 症例報告とは、治験や経験の共有を目的として、希少な病気や重要なメッセージを含む症例についてまとめた報告書である。
% 症例報告には、患者の基礎情報、病歴、主訴、検査結果、診断、治療計画等の情報が含まれる。
% 引用のように、症例の因果関係の理解が重要である。
% A case report contains rich medical entities appearing in patient information, history, findings (e.g., symptoms, test results), and treatment as well as the diagnosis procedure.
A medical case report is a detailed document describing a case involving a rare disease or an important clinical experience, intended to share clinical knowledge.
Each report comprehensively encapsulates the diagnostic process, integrating rich medical entities such as patient information (e.g., age), medical history (e.g., past diseases), clinical findings (e.g., symptoms and test results), and treatments.
As described in \citet{Diagnosis-Still-in-Question}, understanding the causal relationships among medical entities is crucial for comprehending the diagnosis procedure.
In this context, existing NLP research has a history of engaging in medical relation extraction (RE) \cite{parikh-etal-2019-browsing, wolf-etal-2019-term, gao-etal-2023-dialogue-medical, khetan-etal-2022-mimicause} to extract causal relationships between medical entity pairs.
% As described in \citet{Diagnosis-Still-in-Question}, understanding the causal relationships among primary diseases, findings, and treatments is crucial for diagnosis.
% }
% \textcolor{blue}{
% 症例テキスト理解のための因果関係の抽出は診断の手続を理解するのに重要である。
% 医学論文でも因果関係とその整理が症例の理解のために重要であると述べられている。
% 先行研究でも医学テキストについての関係抽出タスクや情報抽出タスクが取り組まれてきた。
% Extracting causal relationships from medical case reports is essential for understanding the diagnostic processes of the cases.
% In this context, existing NLP research has a history of engaging in medical information extraction (IE), including relation extraction (RE) \cite{parikh-etal-2019-browsing, wolf-etal-2019-term, gao-etal-2023-dialogue-medical, khetan-etal-2022-mimicause} to extract causal relationships between medical entity pairs.
% }

% \textcolor{blue}{
% 医師は症例報告を読むことで、診断の根拠となる因果関係を含む、診療に役立つ知見を得ることができる。
% 専門家による診断は、症例の因果木をボトムアップで推論するタスクである。
% 複数の病態を含む医学症例の診断の手続の例を図1に示す。
% 専門家はまず、因果に基づいて情報を整理し、ばらばらの手がかりがそれぞれどのdiseaseに帰属するかを推測する。
% 次に、sub diseaseがどの病態に帰属するかという仮説を立て、subtreeごとに推論を繰り返すことで最終的に第一病態に到達する。
% By reading case reports, experts can gain valuable insights for their practice by understanding the diagnostic procedure of the case.
% By understanding the diagnostic procedure in the case report, clinicians can gain valuable insights for their practice.
% However, beyond pairwise causal relationships,
%causally unexplained findings based on causal relationships. 
% An expert's
Clinicians can gain valuable insights to enhance their practice by understanding the diagnostic procedure of an existing case~\cite{diagnosis-learning}.
The diagnosis procedure is often carried out in a bottom-up manner, resulting in a comprehensive causal tree extracted from the case report.
An illustration of the diagnostic procedure is shown in Figure \ref{fig:diagnosis-flow}.
In this process, first, clinicians organize a list of findings as leaves.
Second, clinicians predict which causative disease corresponds to some set of leaves, and construct a hypothetical causal subtree with causative disease as the parent.
Third, the parents of subtrees serve as children in the bottom-up procedure and clinicians iteratively infer the parent of each subtree.
Finally, clinicians reach the primary disease as the root and derive the most plausible causal tree that can describe the entire case.
This indicates the limitation of the existing RE task for pairwise causal relationships that lacks consideration of multi-layered causal structures.
Consequently, they are insufficient for demonstrating expert-like medical text comprehension procedures.
Therefore, we propose a novel \textbf{causal tree extraction (CTE) task} that transforms case reports into a \textbf{causal tree}.
An example of a causal tree is shown in the top box in Figure \ref{fig:diagnosis-flow}.
The most distinctive characteristic of CTE is that it form a tree structure with the primary diseases as roots.
The causal tree presents an at-a-glance understanding of which parts of the case are important and what the main causal consequences are, even if the reader lacks specialized knowledge.
In addition, causal relations has the potential to enhance the keyword searching capabilities of case report databases.
In this paper, we present a full pipeline of the construction of a human-annotated CTE dataset, LLM-based CTE method, and evaluation metrics.
% S, we subsequently present a generation-based approach to solve including data construction, model training and evaluation for CTE task.
First, we construct the \textbf{J-Casemap} dataset, which consists of Japanese case reports and their corresponding causal trees.
The causal trees in the J-Casemap have been annotated by highly specialized Japanese clinicians, and further experiments show their benefits on medical QA tasks, making them a potential resource for various medical applications.
% }

% \textcolor{blue}{
% 次に、私たちはLLMのテキスト生成ベースの自動生成手法を提案する。
% 近年の商用LLMは医学分野についても高い性能を発揮している。
% しかし、大規模な商用モデルはデータリークの懸念のために患者のデータを入力することが禁じられている。
% それゆえに我々は13B規模のモデルによる実験に取り組むが、モデルサイズが小さくなると、次は医学知識の不足による能力低下が問題視される。
% 我々は医学知識の不足を補うため、日本語医学データによる継続事前学習についても取り組む。
% 実験では、継続事前学習はモデルのCTEタスクの性能を向上させ、特にSFTデータの少ない設定において効果を発揮することが示された。
% また、継続事前学習を含む提案手法によるCTEタスクへの取り組みでは、人手評価スコア82.7を達成し、既存手法を上回った。
% Next, we propose a generation-based prediction method for CTE.
% Though recent LLMs have demonstrated high performance in the medical domain \cite{kasai2023evaluating}, large commercial models like ChatGPT \cite{openai2024gpt4technicalreport}, Claude \cite{Claude2}, Gemini \cite{geminiteam2024gemini15unlockingmultimodal} are restricted from processing patient data due to data leakage concerns.
% Therefore, we conduct experiments using Japanese specialized open LLMs and combine continual pretraining with Japanese medical data and fine-tuning for CTE to compensate for the lack of medical knowledge. 
% % However, as model size decreases, performance degradation becomes a challenge.
% Experimental results show that continual pretraining improves the performance of the model on the CTE task, especially in a low-resource supervised data setting.
% Additionally, the proposed method, which includes continual pretraining, achieved a human evaluation score of 82.7. 
% This score is 20.2 pt higher than existing approaches for the CTE task.
% % 20.2pt上昇
Next, we propose a generation-based method for CTE.
Though recent LLMs have demonstrated high performance in the medical domain \cite{kasai2023evaluating}, large commercial models like ChatGPT \cite{openai2024gpt4technicalreport}, Claude \cite{Claude2}, Gemini \cite{geminiteam2024gemini15unlockingmultimodal} are restricted from processing patient data due to data leakage concerns.
Therefore, we conduct experiments using Japanese specialized open LLMs and combine continual pretraining with Japanese medical data and fine-tuning for CTE to compensate for the lack of medical knowledge. 
The proposed method achieves a human evaluation score of 82.7, which substantially outperforms the baseline~\cite{Ozakimodel} by 20.2 points. 
Ablation study shows the effectiveness of continual pretraining, especially in the low-resource setting.
\textcolor{blue}{
% Additional experiments showed that incorporating J-Casemap along with the medical QA data for SFT could substantially improve the performance of medical QA tasks, which shows the potential of being universally beneficial to other downstream tasks.
}
Finally, we propose an automatic evaluation method that reflects clinician preferences since human evaluation requires highly experienced clinicians and is costly.
% In evaluating the CTE, the important factors are whether the primary disease of the case is correctly placed at the root and whether the primary disease has child diseases that are caused by itself.
In evaluating CTE, the important factors are whether the primary disease of the case is correctly extracted and whether relationships associated with those nodes at the higher layer of the tree are correctly extracted. 
% whether the causal relationships between diseases are correctly extracted and located at the higher layer of trees.
Conversely, the absence of extracted entities that are less related to the diagnosis is not a critical issue.
For such a task, existing automatic evaluation methods, such as triplet F1 used in relation extraction tasks is not suitable because they cannot determine the importance of each entity or its position in the causal tree.
Since this evaluation requires extensive medical knowledge, we propose a method that weights relational triplets and focuses on the salient entities based on human preference.
This weighting method reduces the gap between automatic evaluation scores and manual evaluation scores, improving their correlation.
% }

% \textcolor{blue}{
% We summarize our contributions as follows: 
% (1) 高度な因果推論が必要な要約タスクCTEを提案。
% (2) 高品質な因果木がアノテーションされた症例報告からなるCTEデータセットを構築。
% (3) LLMのテキスト生成ベースのアプローチを用いた因果木の予測モデルを提案。
% (4) 症例報告のCTEタスクのための自動評価手法の提案。
We summarize our contributions as follows:
(1) Introducing a novel CTE task that requires advanced text comprehension and constructing the J-Casemap dataset consisting of case reports annotated with high-quality causal tree annotation;
(2) Proposing an LLM-based generative model for extracting causal trees from case reports;
(3) Discussing an automatic evaluation method for CTE on case reports.

\section{Task Definition: Causal Tree Extraction (CTE)}
\label{sec:cts}
\begin{figure*}[t]
    \centering
    \includegraphics[width=\linewidth]{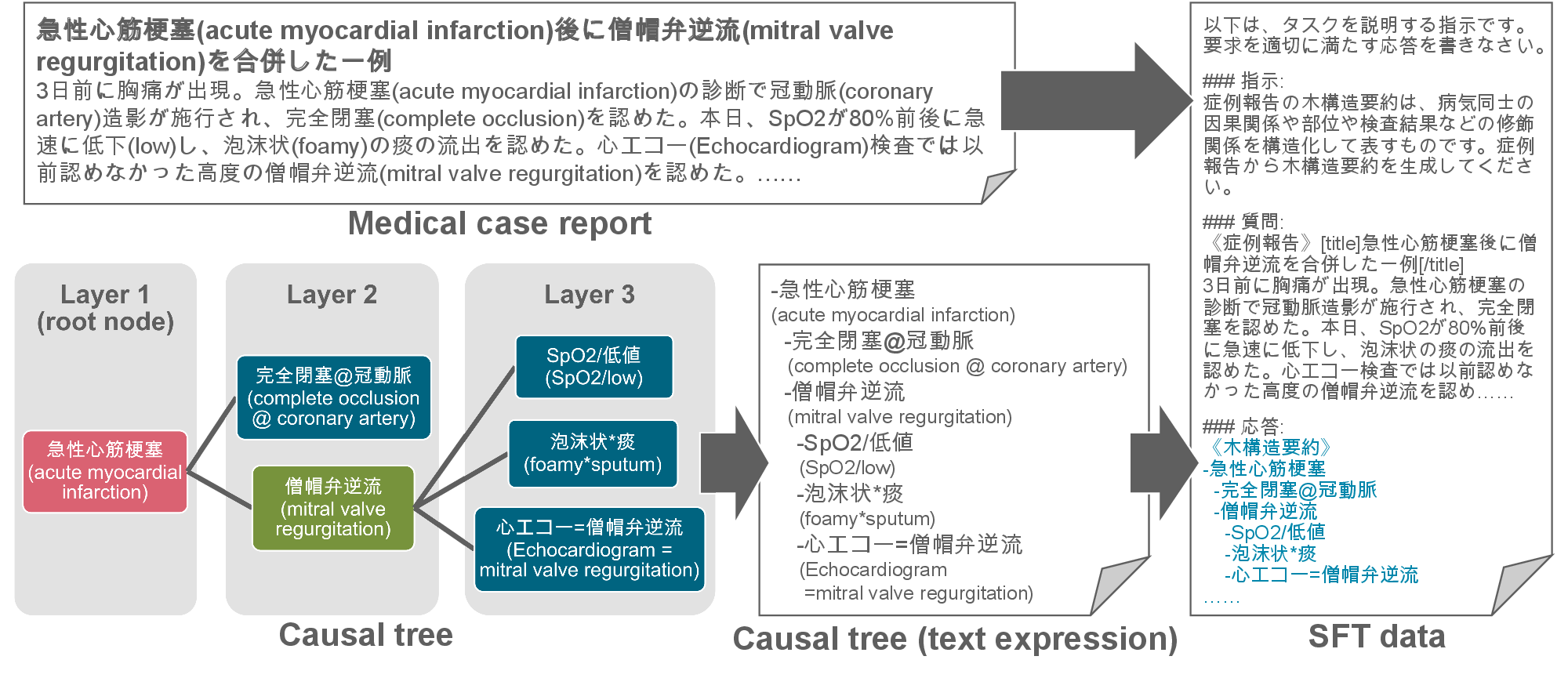}
    \caption{
    % Workflow of the proposed LLM-based method and example of training data prompts. Supervised fine-tuning LLMs for the task of automatic causal tree generation with pairs of abstracts of case reports and manually annotated causal tree.
    The SFT data example.
    }% キャプション
    \label{fig:workflow}
\end{figure*}

% ##### Overall Specifications of Structured Summaries
This section explains the specifications of the CTE task.
A medical case report is represented as a disease-centric tree, where each \textbf{node} offers the modification information surrounding a head entity (usually a disease or finding), and the \textbf{edges} between nodes usually represent the causal or evidential \textbf{\textit{parent\_of relation}} between diseases and findings. For instance,  the root “\begin{CJK}{UTF8}{min}急性心筋梗塞\end{CJK} (acute myocardial infarction)'' is evidenced by the child “\begin{CJK}{UTF8}{min}完全閉塞\end{CJK} (complete occlusion)'' in Figure~\ref{fig:workflow}.
The \textbf{root} node of the tree structure corresponds to the primary disease, which represents the main factor that causes other diseases or findings.
Then, we link those evidential nodes through edges (representing \textit{parant\_of} relationships) to the root. 
These diseases may also cause their own child nodes, naturally extending the \textbf{depth} of a tree summary.
% The length of the path from the root node to a node is called the \textbf{depth} of that node.
% We refer to a group of nodes of the same depth as a \textbf{layer}.

% ##### Modifier Symbols
To be noticed, each node can have internal structures, expressing the supporting information modifying the head entity.
There are four pre-defined modification relationships and corresponding text symbols are denoted as follows:

% \noindent\textbf{Anatomical Location (symbol: @).}
\noindent\textbf{\textit{located relation} (symbol: @):}
	Represents the anatomical location of a disease or finding (e.g., ``\begin{CJK}{UTF8}{min}完全閉塞\end{CJK} (complete occlusion) @ \begin{CJK}{UTF8}{min}冠動脈\end{CJK} (coronary artery)'').

% \noindent\textbf{Polarity Information (symbol: /).}
\noindent\textbf{\textit{polarity relation} (symbol: /):}
	Indicates whether a test result is high or low, or whether a treatment was effective or not (e.g., ``SpO2 / \begin{CJK}{UTF8}{min}低値\end{CJK} (low)'').
        % \textcolor{blue}{
            % 症例報告テキスト中において数値で示されている検査結果は、因果木の中では全て極性に変換される。
            All numerical test results in the case report are converted to polarity within the causal tree.
        % }

% \noindent\textbf{Specimen Name / Test Name (symbol: =).}
\noindent\textbf{\textit{tested relation} (symbol: =):}
	Specifies the test from which a finding was obtained (e.g., ``\begin{CJK}{UTF8}{min}心エコー\end{CJK} (Echocardiogram) = \begin{CJK}{UTF8}{min}僧帽弁逆流\end{CJK} (mitral valve regurgitation)'').

% \noindent\textbf{Supplementary Information (symbol:＊).}
\noindent\textbf{\textit{featured relation} (symbol: \begin{CJK}{UTF8}{min}＊\end{CJK}):}
	Represents details such as laterality or appearance features of a disease or finding (e.g., ``\begin{CJK}{UTF8}{min}泡沫状\end{CJK} (foamy) \begin{CJK}{UTF8}{min}＊\end{CJK} \begin{CJK}{UTF8}{min}痰\end{CJK} (sputum)'').
 
The head entity of located or polarity relation is the preceding one and that of tested and featured relation is the succeeding one.
Modifier relationships can be combined, such as in ``MRI = DWI\begin{CJK}{UTF8}{min}高信号\end{CJK} (high signal) @ \begin{CJK}{UTF8}{min}右\end{CJK} (right) \begin{CJK}{UTF8}{min}＊\end{CJK} \begin{CJK}{UTF8}{min}大脳半球\end{CJK} (cerebral hemisphere).''
For example, the case in Figure~\ref{fig:workflow} shows that the condition of acute myocardial infarction caused chest pain, complete coronary artery occlusion, and mitral valve regurgitation.
Moreover, ``mitral valve regurgitation'' resulted in a ``low SpO2'' test result and ``foamy sputum'', and it was observed through an ``echocardiogram''.

% \textcolor{blue}{
    % さらに、主辞ノードは特殊接頭記号\textbf{H:}を持つ場合がある。
    % この記号はそのノードが既往歴または治療歴であることを示す。
    % 例えば、"H:アルコ-ル性肝線維症"はその親の病態の背景に"H:アルコ-ル性肝線維症"という既往歴があることを示す。また、"H:ステロイド/有効"は、その親エンティティに対しステロイドを投与した結果が有効であったことを示す。
    In addition, the head node may have a special prefix, \textbf{H:}.
    This symbol indicates that the node represents a medical history or treatment.
    For example, "H:\begin{CJK}{UTF8}{min}アルコール性肝線維症\end{CJK} (Alcoholic liver fibrosis)" indicates that the parent disease has a history of alcoholic liver fibrosis.
    Similarly, "H:\begin{CJK}{UTF8}{min}ステロイド\end{CJK} (Steroid) / \begin{CJK}{UTF8}{min}有効\end{CJK} (Effective)" indicates that the parent entity was treated with steroids, and the treatment was effective.
% }

% % \textcolor{blue}{
%     % 診断能力の獲得のためには、疾患とその症状の因果関係に関する医学知識と実際の臨床経験から得られる知識の両方が重要であると考えられる。
%     % \cite{}によると、医学の初学者は患者の主訴から候補となる病態を列挙することはできるが、優先順位をつけられない傾向にあると考えられる。
%     % その一方で、熟練者は症例を整理・抽象化することで要点を把握し、候補となる病態と他の要素の因果関係についての仮説を立てる。
%     % さらに、熟練者は知識や経験を活かして仮説の比較や情報の絞り込みを行い、効率的に鑑別名を推論する。
%     The bottom up interpretation of causal tree is like a expert's diagnotic procedure.
%     As discussed in \cite{diagnosis-learning}, novice clinicians often take numerous diseases as candicates but struggle to select the most plausible one.
%     In the other hand, an expert organizes the case, then formulates hypotheses based on the causal relationships between candidate diseases and other entities.
%     Furthermore, an expert leverages knowledge and experience to  efficiently select plausible hypothesis.
% % }

% % \textcolor{blue}{
%     % 臨床診断は、因果の木の下流から上流を求めるボトムアップ方式のタスクと見なすことができる。
%     % 因果の木は、トップダウンで解釈することで症例全体の因果の流れを理解できるだけでなく、ボトムアップで解釈することで初学者でも医師の診断の流れを感じ取ることができる。
%     % CTEタスクによる因果の木は、医学LLMの訓練のためのデータとしても有用な可能性を持つ。
%     % さらに、明示された関係を利用することで、症例データベースの検索性を拡張するポテンシャルを持つ。
%     A causal tree can provide even beginers not only bottom-up understanding of a diagnostic reasoning process but also top-down understanding the causal flow of a case.
%     A causal tree generated from the CTE task has the potential to serve as valuable training data for medical LLMs.
%     In additional, causal trees extend the search capability of case report database with the explicitly expressed relationship.
% }

% ##### triplets

\begin{figure}[t]
    \centering
    \includegraphics[width=\linewidth]{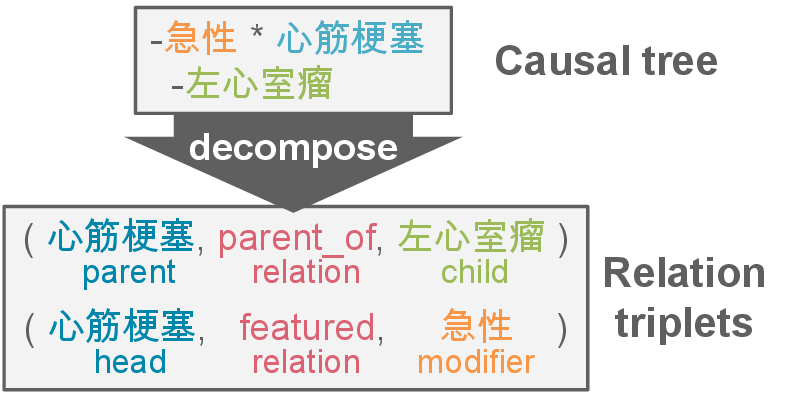}
    \caption{A tree summary is decomposed into triplets.}% キャプション
    \label{fig:Triplet}
\end{figure}

% ##### Potential of Structured Summaries in Case Reports
\subsection{Dataset Construction: J-Casemap}
This subsection introduces the collection of the CTE dataset, named J-Casemap.
All annotated data are based on case reports in internal medicine.
%For training, we used a dataset comprising 15,100 pairs of case report abstracts and their corresponding human-annotated causal trees.
% Among these cases, some contained multiple case reports in a single entry, or included only URLs pointing to PDF files instead of actual case text, making them unsuitable for the current experiment.
% The doctor who annotated the data is very skilled and the quality of the annotations is very high.
The most experienced doctor (a co-author of this paper) first drafted the annotation schema.
The annotation was then conducted by the doctors with at least ten years of experience.
They made iterative revisions to the annotation schema and cross-validation of the annotation for years to complete around 15,000 medical case reports.
After excluding inappropriate data, the final dataset consisted of 14,094 cases.
% annotation quality
% Motivation of automatic generation

% There is a need to speed up the annotation of structural summaries.
% \textcolor{blue}{
    % J-Casemapデータセットに含まれる症例報告およびcausal treeの全ては、
    % 症例検索データベースJ-CaseMap~\footnote{anonymous link}において日本内科学会の会員向けに公開されている。
    All case reports and causal trees included in the J-Casemap dataset are available in the J-CaseMap case search database~\footnote{https://www.naika.or.jp/j-casemap/}, exclusively for members of the Japanese Society of Internal Medicine.
% }
Although the actual J-Casemap dataset and the statistics for a close look at annotation history are not allowed for publication, we will release the final version of the annotation schema and 100 causal tree samples based on public case reports from the Japan national medical license examination.~\footnote{preparation in progress.}

\section{Automated CTE Models}
This section introduces two comparable methods of automatic causal tree generation: the RE method and the generation method.

\subsection{RE Method (baseline)}
\label{sec:RE}
% % tripletはgeneration modelでも使うので上に移動してもいい？
% For RE methods, CTE can be represented as a set of \textbf{triplets} comprising two entities and the relationship between them~\cite{Ozakimodel}.
% Modifier relationships within a node are expressed as triplets consisting of the head entity, the modifying entity, and the modification relationship.
% Parent-child relationships between nodes are expressed as triplets consisting of the head entity of the parent node, the head entity of the child node, and the parent-child relationship.

The RE task is originally designed to extract triplets of relationships between entities, instead of the tree structure. 
Thus, we first decompose a tree summary into a list of triplets (Figure \ref{fig:Triplet}) with each triplet assigned by one relation type among the set: \{\textit{parent\_of}, \textit{located}, \textit{polarity}, \textit{tested}, and \textit{featured}\} defined in Section~\ref{sec:cts}.

RE methods typically require the span information of entities in case reports to perform the initial named entity extraction. 
However, the span position of a node inside the case report is not provided in our causal tree extraction task. 
\citet{Ozakimodel} used the idea of distant supervision to flexibly match a node with the words in the case report to determine its span and generate pseudo data. 
They trained a supervised model on this pseudo data to predict triplets. 
We follow their method to train an RE model, which serves as the baseline of this paper.
It is evident that such distant supervision inevitably generates a significant amount of noise when matching entity spans, which becomes a bottleneck to limit the performance of RE models.

Recently, generation-based approaches~\cite{zeng2020copymtl,zhang-etal-2020-minimize,wadhwa-etal-2023-revisiting,wan2023gptre} in an end-to-end manner (i.e., shorten the need of span information) have achieved performance on sentence-level RE tasks that rivals or even surpasses traditional RE models. Moreover, the fact that LLMs have recently passed the Japanese medical licensing exam~\cite{kasai2023evaluating}, suggests LLMs are capable of learning extensive medical knowledge. All these findings indicate that the LLM-based generation method could be highly suitable for our CTE task. The potential challenge lies in that our task is much more complex than sentence-level RE.

\subsection{Generation Method (proposal)}

% ##### Definition of the Generation Model
In this study, we propose to solve CTE using LLMs, referred to as the generation model. Apart from not relying on noisy spans like RE models, the generation model also benefits from being able to refer to previously predicted triplets as context, allowing it to maintain consistency across triplets.

% ##### Fine-Tuning Workflow and Advantages of the Approach
% The generation workflow is shown in Figure \ref{fig:workflow}.
% ##### Conversion from Graph Representation to Text Representation of Structured Summaries
Since LLMs take textual input of the pairs of case reports and tree summaries, the tree structure must be converted into certain forms of \textbf{text representation} as shown in Figure~\ref{fig:workflow}.
% In this study, the tree structure is represented using indentation .
We converted the tree structure into text using a depth-first linearization method with indentation indicating the depth information. 
% An example of the text representation is shown in Figure \ref{fig:workflow}.
In this representation, each line corresponds to a node, and the depth of indentation indicates the \textbf{\textit{parent\_of}} relationship between nodes.
As recent LLMs are typically trained on datasets that include code (such as Python), using indentation to represent nested structures is considered a natural format for LLMs.
% \textcolor{blue}{
    % また、テキスト表現の決定においてはbracketを利用して入れ子構造を表現する形式を試したが、ベースモデルによっては入れ子構造が壊れて評価不能となったため、今回は採用しなかった。
    For determine the textual representation of the tree structure, we also experimented with a bracket-based format to represent the nested structure. However, it was not adopted because the nested structure broke down the output format, making evaluation impossible.
% }

%Furthermore, since this method does not require annotations indicating which spans in the input text correspond to the entities, it avoids the bottleneck of weakly supervised data quality.
% SFT method
We conduct two-step training to derive our generation model.
\paragraph{Continual pretraining:}
Since solving CTE requires highly specialized expertise, we leverage continual pretraining to inject the Japanese medical domain knowledge into the base models.
Our Japanese medical corpora are collected from two sources. One is the abstracts from Japanese medical papers, the other one is the Japanese version of English MedPub translated by human experts.
In summary, we collect high-quality medical data for the pretraining process.

\paragraph{Supervised fine-tuning:}
We implemented supervised fine-tuning on our collected J-Casemap data as shown in Figure \ref{fig:workflow}.
Supervised fine-tuning (SFT) is a technique that uses labeled data to adapt pre-trained LLMs to specific downstream tasks.
The prompt template filled with pairs of case reports and tree summaries is fed into LLM for SFT. The blue part in the prompt demonstrates that only the tree summary is used to calculate the cross-entropy loss for updating model parameters.

% Due to differences in model compatibility, two types of inference templates were used according to the model.
% The inference templates follow the examples provided on the model card for each model.
% In this experiment, the blue parts of the figure \ref{fig:workflow} were used for loss calculation.
% Cross entropy loss was used as the loss function.

% Lora, promptの細かい話はexperiment settingへ

%\section{causal tree extraction Task}

% ##### Need for Automatic Generation Task
%Creating structured summaries requires advanced medical expertise to understand the causal relationships between pathologies and findings, as well as a deep understanding of annotation specifications, such as parent-child relationships and modifier symbols.
%This process is typically limited to medical professionals, making it time-consuming and difficult to perform at a high speed.
%To address this issue, automating the generation of structured summaries from case reports, followed by manual post-editing, could accelerate the annotation process.
%This section explains two approaches to automatic generation: RE models and generation models.

\section{Evaluation}
% ##### Challenges and Overview of Evaluation Methods
%Existing automatic evaluation methods, either triplet F1 commonly used in RE tasks or ROUGE in summarization tasks, fails to differentiate the importance of nodes in a tree summary (e.g., a root node typically representing the primary disease intuitively should be more important than those minor leaves)
Comprehensively evaluating CTE requires the medical perspective of human clinicians to differentiate the importance of nodes, \textit{parent\_of} relationships, and modifiers for extracting salient diagnostic information. 
Since existing automatic evaluation metrics in RE fail to align with human clinicians (as later shown in section \ref{subsec:comparison weight}), we propose a weighting method emphasizing human preference to narrow the gap.
%considering not only entity extraction accuracy but also the depth of parent-child relationships and modifier relations. is a highly complex and difficult task, since existing automatic methods fails to align with 
%as it requires considering not only entity extraction accuracy but also the depth of parent-child relationships and modifier relations. 
%In this paper, two methods were used to evaluate the quality of structured summaries: manual evaluation by clinicians and automatic evaluation.

In this section, we will introduce the manual evaluation and the automatic evaluation, including our proposed weighting method. 
%Later in section \ref{subsec:comparison weight}, we compare automatic and manual evaluation as a preliminary experiment, and the results show the better alignment of our weighting method to manual results.

\subsection{Manual Evaluation} \label{subsec:manual evaluaction}
% Evaluation Criteria and Dataset Size

% The manual evaluation was scored on a scale of 0 to 100.
% Scoring criteria is based on the amount of manual post-edit needed.
% If the focus on the case is correct, the causal tree achieve at least 60 points.

The manual evaluation is scored on a scale of 0 to 100.
The scoring criteria mainly follow a deduction system, where the less amount of manual post-edit is needed, the higher score is assessed, and vice versa.
% ##### Characteristics of Manual Evaluation: Emphasis on the Overall Structure
% 大筋の日本語訳おかしいかも
% In manual evaluation by clinicians, \textbf{the focus} of the case report is emphasized.
Human doctors are naturally allowed to focus more on those important diseases and any associated diseases in the tree structure based on their expertise.
% Consequently, modifier relations such as findings and locations are considered less important than the focus on the tree.
Consequently, modifier relations such as findings and locations are considered less important than the \textit{parent\_of} causal relations in the trees.
The most important disease often corresponds to the root node of the tree, and shallow layers tend to be more important than deeper layers.

\subsection{Automatic Evaluation}

% ##### Evaluation Based on the F-Score of triplets
To utilize automatic metrics, the output of the structured summary was broken down into a set of triplets, which were then compared to the set of correct triples.
A correct prediction was defined as one where both the entities and the relationship between them matched.
Precision, Recall, and F-score were calculated based on the number of correct prediction triplets.

% ##### Entity Matching Criteria
In a entity matching for judging the correctness of triplet, minor variations in notation and typographical errors were allowed to some extent.
First, a thesaurus was used to convert entities into their representative forms.
Next, the edit distance between the output and correct entities was divided by the length of the correct entity, and if this ratio was below a threshold, the entities were considered a match.
In this experiment, the threshold was empirically set at 0.5.
However, for polarity information among modifier relations, no variations were allowed, and only exact matches were considered correct.

% ##### Details of triplet Matching
%For triplet matching, a triple was considered correct if both the pair of entities and the relationship between them matched.
% 後処理の話
% In addition, if the same parent node produced multiple identical childless nodes, these were merged before evaluation.
% When multiple matching entities were found, only the best match was counted, and any triplets that had already been matched were excluded from further matching.

\begin{figure*}[t]
\centering
\includegraphics[width =0.7\linewidth]{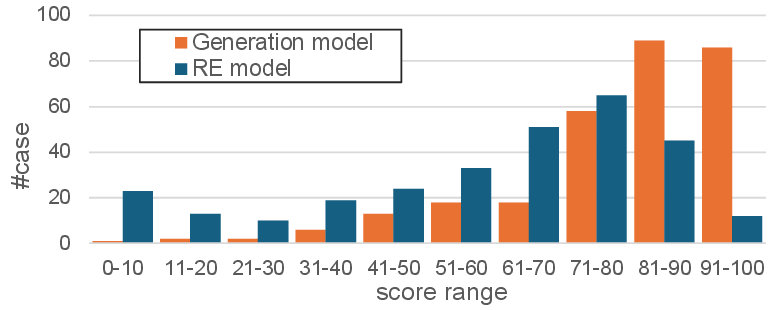}
\caption{Manual evaluation of the generation and RE models on the same 300 cases. 
The generation model achieved an average score of $82.7$, and the RE model achieved an average score of $62.5$.}\label{fig:manual-evaluation}
\end{figure*}

\begin{table*}[t]
\centering
\begin{tabular}{lrllllll}
\hline
                              & \multicolumn{6}{c}{Triplet-based evaluation}                                                                                                                          &  \\
                              & \multicolumn{3}{c}{w/o weight}                                                   & \multicolumn{3}{c}{w/ weight}                                                         &   \multicolumn{1}{c}{Manual evaluaton}            \\ 
                              & \multicolumn{1}{c}{P} & \multicolumn{1}{c}{R}     & \multicolumn{1}{c}{F1}    & \multicolumn{1}{c}{P}     & \multicolumn{1}{c}{R}     & \multicolumn{1}{c}{F1}    &                           \\ \hline
RE model (DeBERTa)            & 50.7                 & \multicolumn{1}{r}{48.2} & \multicolumn{1}{r}{49.4} & \multicolumn{1}{r}{41.2} & \multicolumn{1}{r}{51.4} & \multicolumn{1}{r}{45.8} & \multicolumn{1}{r}{62.5} \\
LLM-jp-13b-v1 & 48.0                 & \multicolumn{1}{r}{48.9} & \multicolumn{1}{r}{48.4} & \multicolumn{1}{r}{50.5} & \multicolumn{1}{r}{50.0} & \multicolumn{1}{r}{50.2} & \multicolumn{1}{r}{82.7} \\ \hline
\end{tabular}
\caption{ The comparison between automatic and manual evaluation on the subset of 300 test cases. 
To be noticed, manual scores ranging from 0-100 are not directly comparable to the automatic triplet F1.
% The gold data for automatic evaluation is created by modifying the results of the generated model.
}\label{tab:manual-evaluation}
\end{table*}

% ##### Weighting of triplets
\paragraph{Proposed weighting method}
Since existing triplet-based evaluation treats all triplets evenly, it fails to reflect human preference.
In our experiments, each triplet was weighted based on the depth $ d $ of the node and the presence of modifier relations.
% tripletのd計算方法
% The depth of an entity is calculated as the $1+$depth of it's parent entity, and the depth of ``Triplet(arg1, relation, arg2)'' is equal to the depth of arg1.
The depth of an entity is calculated as the depth of its parent entity plus $1$, and the depth of a triplet is equal to the depth of the parent entity or the head entity inside.
In our automatic evaluation method, when decomposing causal trees into triplets, we use a dummy entity ``[root]'' with the depth $d=0$ as the parent of the root node.
For the example in Figure~\ref{fig:workflow}, the depth of the triplet ``([root], parent\_of, \begin{CJK}{UTF8}{min}急性心筋梗塞\end{CJK})'' is $0$, and the depth of the triplet ``(\begin{CJK}{UTF8}{min}急性心筋梗塞\end{CJK}, parent\_of, \begin{CJK}{UTF8}{min}僧帽弁逆流\end{CJK})'' is $1$.
We design a weighting method of each triplet as follows:

\noindent Weighting method: $$ W = \frac{1}{1+Cd}x_{relation} $$
$ x_{relation} $ is $ 1 $ when the relation type is $ parent\_of, $ and $\frac{1}{2}$ if not.
$C$ is a constant hyper-parameter that can be tuned. $d$ is the triplet depth. 

These weighting methods are heuristically determined by referencing the manual evaluations conducted by highly experienced clinicians, who emphasized those top layers in the tree summaries (e.g., the root) and \textit{parent\_of} relations over other relation types.
% \textcolor{blue}{
    % 重みの計算式およびハイパーパラメータ決定の詳細についてはAppendix\ref{sec:appendix-triplet-weight}に示す。
    Details of the weighting formula design and hyperparameter selection are provided in Appendix \ref{sec:appendix-triplet-weight}.
% }
The hyperparameter $C=2$, which shows the highest correlation coefficients to human scores, is used in all the following experiments.

\section{Experiment Setups}

This section describes the settings for continually pretraining and fine-tuning our LLM for CTE. More Details like hyper-parameters are shown in Appendix~\ref{appendix:hyper}.

\noindent\textbf{Base LLMs}
As general-domain LLMs for Japanese processing, we leverage the instruct version of multilingual Japanese LLM-jp-13b-v1~\cite{DBLP:journals/corr/abs-2407-03963}, and Japanese Swallow-13b~\cite{Fujii:COLM2024}. 

\noindent\textbf{Continually pretraining}
We conducted on two nodes of 8 x NVIDIA A100 40GB and totally trained one epoch on the $2$B tokens for each model. 
For those continually pre-trained LLMs, we re-name them by adding the suffix ``Med-."

\noindent\textbf{Supervised fine-tuning}
We divided J-Casemap into 13,426 training cases, 200 development cases, and 468 test cases.
Prompt templates are shown in Appendix~\ref{appendix:LLMprompt}.
We used LoRA~\cite{hu2021lora} as the SFT method.

\noindent\textbf{Baseline RE Model}
As a baseline, we fine-tune models via the distant supervision approach mentioned in Section~\ref{sec:RE}. JaMIE~\citep{cheng-etal-2022-jamie} is the backbone RE model, and the encoder is initialized by Japanese DeBERTa~\cite{he2021debertav3}.

\section{Experimental Results}

% ##### Comparison Between Automatic and Manual Evaluation Methods
\subsection{Pre-examination for Optimizing Automatic Evaluation}\label{subsec:comparison weight}

As a pre-examination of evaluation metrics, we chose the RE model and generation models based on LLM-jp-13b-v1 as our subjects.
We fine-tune both models on the J-Casemap train set.
We randomly sample 300 cases from the test set to compare the automatic and manual evaluations for the RE and generation models. 
To be clarified, the manual evaluation is scored on a scale of 0-100 and is not directly comparable to the automatic F1 score.
% \textbf{Analyzing the gap between automatic and manual evaluation:}
Figure \ref{fig:manual-evaluation} shows the manual evaluation results. 
The generation model achieved an average score of $82.7$, significantly outperforming the RE model by 20.2 points.

% In Table \ref{tab:manual-evaluation}, the automatic evaluation without weight adjustment shows very close scores between the two models, which is far from human preferences.
%As we expected, automatic evaluation, which does not find the focus on the tree, deviated from the manual score. 
% \textcolor{blue}{
% 表\ref{tab:manual-evaluation}は重みづけの有無による自動評価スコアの違いを示す。
% 関係三つ組の重み付けがない場合、人手評価スコアと自動評価スコアとの間の相関係数は0.604であった。
% また、GenerationモデルよりもREモデルの方が高い自動評価スコアを獲得した。
% この結果は人手評価の結果と逆転している。
% Table \ref{tab:manual-evaluation} shows the changes in scores with and without triplet weighting.
% In the case without weighting, the correlation between the automatic score and the human score is 0.604.
% While this correlation is relatively strong, The RE model received a slightly higher score than the generation model, which contradicts the human evaluation results.
% This inconsistency suggests that the automatic evaluation, lacking a focus on the root of the tree, does not align with human evaluation.
However, in the vanilla triplet evaluation (w/o weight) of Table~\ref{tab:manual-evaluation}, the RE model obtained a slightly higher score than the RE model, which substantially contradicts the human evaluation results.
Such inconsistency suggests that the vanilla metric, lacking a focus on those salient entities, does not align with human evaluation.
% }
% \textcolor{blue}{
% 一方、関係三つ組の重み付けがある場合、人手評価スコアと自動評価スコアとの間の相関係数は0.646と改善した。
% また、REモデルよりもGenerationモデルの方が高い自動評価スコアを獲得した。
% このことから、重み付けは人間の医師の好みを評価に反映していると見なすことができる。
After the weighting method was applied, the correlation between the triplet score and the human score was improved from 0.604 to 0.646 in Figure~\ref{fig:correlation}. Consequently, the generation model obtained significantly higher scores than the RE model in the new metric (w/ weight), which suggested improved consistency with the human evaluation and better reflection of the doctors' preferences.

\begin{table*}[]
\centering
\begin{tabular}{cllrrr}
\hline
\multicolumn{2}{c}{}                                                                                                            & \multicolumn{1}{c}{Domain} & \multicolumn{1}{c}{Precision} & \multicolumn{1}{c}{Recall} & \multicolumn{1}{c}{F1} \\ \hline
\begin{tabular}[c]{@{}c@{}}RE model \cite{Ozakimodel}\end{tabular}                             & DeBERTa                       & general                    & 40.7                         & 50.1                      & 44.9                   \\ \hline
\multirow{4}{*}{\begin{tabular}[c]{@{}c@{}}Generation model\\ (Proposed   method)\end{tabular}} & LLM-jp-13b-v1       & general                    &    48.2                           &       49.1                     &     48.6                    \\
                                                                                                & Swallow-13b             & general                    & 52.0                         & 54.2                      & 53.3                   \\ \cline{2-6} 
                                                                                                & Med-llm-jp-13b-v1 & medical                    &   48.3                 & 49.2                           &        48.8                 \\
                                                                                                & Med-swallow-13b     & medical                    & \textbf{52.8}                & \textbf{54.3}             & \textbf{53.6}          \\ \hline
\end{tabular}
\caption{The overall scores of automatic evaluation for the CTE task. 
% The scores of the generated models outperformed those of the RE models for all base models.
}\label{tab:automatic-evaluation}
\end{table*}

\begin{figure}[t]
    \includegraphics[width=1.0\linewidth]{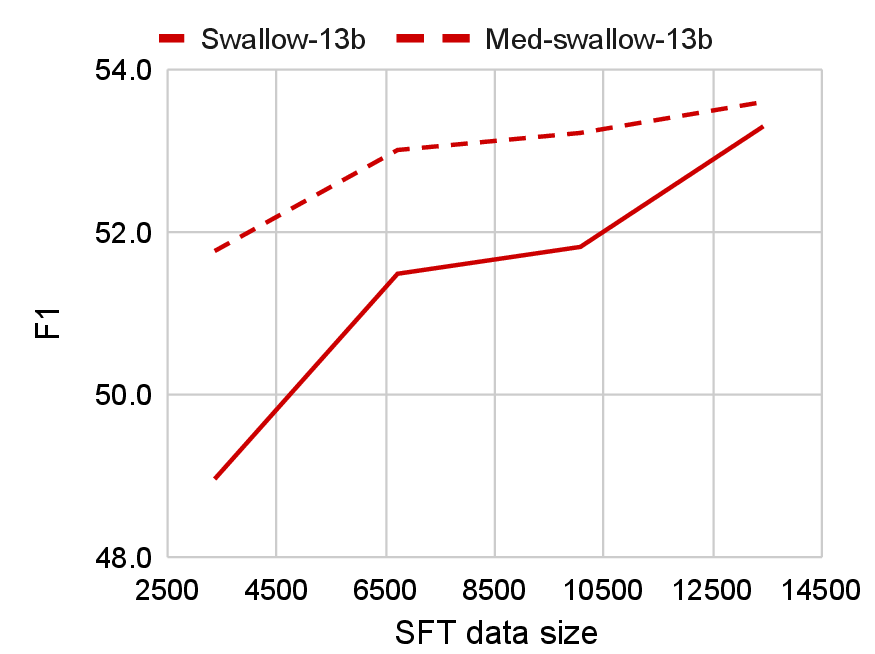}
    \caption{
        % \textcolor{blue}{
        Triplet-based F1 scores of fine-tuned models in settings with varying amounts of SFT data (100\%, 75\%, 50\%, and 25\%).
        % }
    }% キャプション
    \label{fig:SFTdatasize}
\end{figure}

\begin{table*}[]
\centering
\begin{tabular}{cllrrr}
\hline
\multicolumn{2}{c}{}                                                                                                            & \multicolumn{1}{c}{Domain} & \multicolumn{1}{c}{Precision} & \multicolumn{1}{c}{Recall} & \multicolumn{1}{c}{F1} \\ \hline
\begin{tabular}[c]{@{}c@{}}RE model \cite{Ozakimodel}\end{tabular}                             & DeBERTa                       & general                    & 23.5                         & 67.7                      & 34.9                  \\ \hline
\multirow{4}{*}{\begin{tabular}[c]{@{}c@{}}Generation model\\ (Proposed   method)\end{tabular}} & LLM-jp-13b-v1       & general                    &                64.9               &     59.4                       &        62.0                \\ 
                                                                                                & Swallow-13b         & general                    & \textbf{69.2}                & 63.6                      & \textbf{66.3}         \\ \cline{2-6} 
                                                                                                & Med-llm-jp-13b-v1   & medical                    &     64.9 &60.3 &62.5 \\
                                                                                                & Med-swallow-13b     & medical                    & 66.1                         & \textbf{65.8}             & 66.0                  \\ \hline
\end{tabular}
\caption{The automatic evaluation scores for the root node only. }\label{tab:root-evaluation}
\end{table*}

\subsection{Main Results}

The automatic evaluation scores are shown in Table \ref{tab:automatic-evaluation}.
The scores of the generation models outperformed those of the RE models for all base models.
Compared to the RE model, the generation model produced more entities and triplets of relationships.
This corresponds to the experimental results, which showed a significant difference in precision and a minor difference in recall.

% Surprisingly, Llama-2-13b outperformed the LLM-jp-13b-v1 score despite being an English-centric model.
% The results showed that the multilingual capability of Llama-2-13b, which is pre-trained on a large amount of data, is relatively satisfying.
% Swallow-13b, which outperformed Llama-2-13b, is based on Llama-2-13b with continually pretraining on Japanese data.
% This result shows that language specific training improve Llama-2's capability on this Japanese task.
% The difference in scores between the general and medical models was minor.
% This suggests that the effect was not sufficiently apparent because the amount of data used in the continual pretraining of the medical model was small.
% The medical models universally outperform the general models slightly. 
% This suggests the effectiveness of continually pretraining medical data. But only 2B tokens seem to be not sufficient.
% \textcolor{blue}{
    % 図\ref{fig:SFTdatasize}にデータサイズを変化させた場合ののF1スコアを示す。
    % 医学継続事前学習はSFTデータサイズが小さい場合より効果を発揮することが示された。
    % また、15,000規模のデータセットがあれば、医学継続事前学習がなくとも十分高い品質の因果木予測が可能であることが示された。
    F1 scores of models fine-tuned with different data sizes (100\%, 75\%, 50\%, and 25\%) are shown in Figure \ref{fig:SFTdatasize}.
    The results show that continual pretraining is more effective in low-resource settings for SFT data.
    Additionally, the results suggest that the SFT dataset of 14,000 cases is sufficient to train LLMs for CTE without medical continual pretraining.
% }

As discussed in Section \ref{subsec:manual evaluaction}, manual evaluations by highly experienced clinicians prioritize the salient information on the case.
We compute the automatic evaluation scores for the root nodes only, which can be viewed as a primary disease classification task requiring models to capture the primary disease of a case report, as shown in Table \ref{tab:root-evaluation}.
The precision of the generation model significantly outperformed those of the RE model.
This indicates that the generation model adequately detects the focus of the case compared to the RE model.
Root scores detail is discussed in Appendix \ref{appendix:comparison-methods}.

\begin{figure*}[t]
\includegraphics[width = 1.0\linewidth]{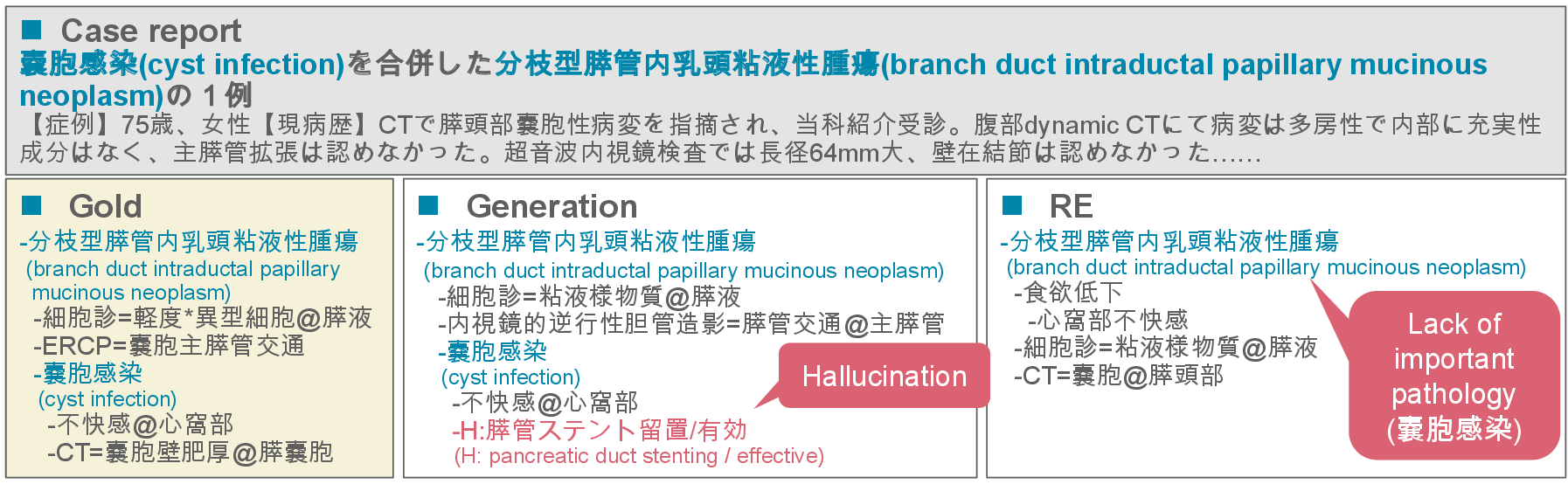}
\centering
\caption{Case study of an automatically generated causal tree. Blue entities are the focus of the tree.}\label{fig:casestudy}
\end{figure*}

% \subsection{Out of Domain Evaluation: Medical QA}
\subsection{Can CTE help Medical QA?}

The J-Casemap data has the potential to serve a variety of other medical tasks, given the comprehensive understanding required for a model to complete the CTE task. 

We conduct the experiments on Japanese medical question answering (QA) benchmarks, like Japanese medical licensing exam dataset IgakuQA~\cite{kasai2023evaluating} and the translated medical QA datasets MedQA~\cite{jin2020disease}, MedMCQA~\cite{pal2022medmcqalargescalemultisubject} to see whether a model trained on J-Casemap can be beneficial to medical QA tasks.
For each benchmark, we used Med-swallow-13b as the base model, and the training set of MedQA or added J-Casemap for fine-tuning; a prompt example is shown in Appendix \ref{appendix:MedQA prompt}.
The evaluation tools were conducted using JmedBench~\cite{jiang2024jmedbenchbenchmarkevaluatingjapanese}.

As shown in Table~\ref{tab:MedQA}, for MedQA, both the ``2-stage'' and ``mix'' settings outperform fine-tuning on MedQA alone. 
For MedMCQA, even fine-tuning on MedQA hurts the performance due to the out-of-domain distribution; after adding J-Casemap in the ``mix,'' the performance improves and beats the base model.
In particular, ``mix'' performs better than ``2-stage'' and achieves the highest scores on all QA datasets. 
This indicates that our J-Casemap data is valuable for facilitating LLMs' medical abilities in various tasks.

% ############# before grammarly #####################
% % ##### Settings
% %We evaluated the accuracy of the QA task using models trained on either the causal tree extraction task, MedQA, or both.
% %A prompt example is shown in Figure \ref{tab:MedQA}.
% % Four types of SFT were conducted: fine tuning using both CTE training data and MedQA training data, fine tuning in CTE followed by fine tuning in MedQA (2-stage), and fine tuning using both CTE and MedQA data at the same time (mix). 
% %Med-swallow-13b was used as the base model.
% % The evaluation tools was conducted using JmedBench~\cite{jiang2024jmedbenchbenchmarkevaluatingjapanese}.
% % MedQA, MedMCQA, and IgakuQA scores are calculated for evaluation.
% % baseよりもよくなった

\begin{table}[]
\begin{tabular}{lrrr}
\hline
\multicolumn{2}{r}{                    MedQA}      & MedMCQA    & IgakuQA        \\ \hline
% \multirow{5}{*}{\begin{tabular}[c]{@{}c@{}}LLM-jp-\\13b-v3\end{tabular}}   & base       & 0.230          & 0.274          & 0.264          \\
%                                  & CTE        & 0.231          & 0.276          & 0.262          \\
%                                  & MedQA          & \textbf{0.330} & 0.315          & \textbf{0.398} \\
%                                  & 2stage & 0.296          & 0.306          & 0.367          \\
%                                  & mix & 0.326          & \textbf{0.318} & 0.377          \\ \hline
% \multirow{5}{*}{\begin{tabular}[c]{@{}c@{}}swallow\\-13b\end{tabular}}     & base       & 0.222          & 0.299          & 0.259          \\
%                                  & CTE        & 0.203          & 0.276          & 0.241          \\
%                                  & MedQA          & \textbf{0.340} & \textbf{0.317} & \textbf{0.356} \\
%                                  & 2stage & 0.320          & 0.307          & 0.332          \\
%                                  & mix & 0.334          & 0.311          & 0.350          \\ \hline
%\multirow{5}{*}{\begin{tabular}[c]{@{}c@{}c@{}}Med-\\swallow\\-13b\end{tabular}} & 
base       & 25.6          & 33.6          & 33.9          \\
                                 % & 
+ J-Casemap        & 22.7          & 29.3          & 26.3          \\
                                 % & 
+ MedQA          & 29.3          & 27.6          & 37.6          \\
                            % & 
+ 2-stage & 34.7          & 32.2          & 34.1          \\
                                 % & 
+ mix   & \textbf{37.0} & \textbf{34.1} & \textbf{38.6} \\ \hline
\end{tabular}
\caption{QA Task Evaluation (Accuracy). We compare the base model with three fine-tuning settings: (1) only J-Casemap; (2)  only MedQA; (3) first J-Casemap then MedQA (2-stage); (4) merge J-Casemap and MedQA data (mix). }\label{tab:MedQA}
\end{table}

\subsection{Case Study}
Examples of causal trees generated by the generation model are shown in Figure~\ref{fig:casestudy}.
Most errors in the generation model's output are failures of entity extraction.
Additionally, the problem of hallucinations, where the model generates entities not present in the original case report, was sometimes observed in the causal trees.
In contrast, due to the nature of information extraction, RE models did not exhibit such hallucinations.
Further studies will explore to what extent the hallucination issue can be mitigated through improvements to the base LLM or additional training using medical domain texts.

% ############# before grammarly #####################
% % ##### Case Study and Hallucination
% Examples of causal trees generated by the generation model are shown in Figure~\ref{fig:casestudy}.
% Most errors in the generation model's output are entity extraction errors.
% Additionally, the problem of hallucinations, where the model generates entities not present in the original case report, was observed in some outputs.
% In contrast, due to the nature of information extraction, RE models did not exhibit such hallucinations.
% Further studies will explore to what extent the hallucination issue can be mitigated through improvements to the base LLM or additional training using medical domain texts.

\section{Related Works}

% \subsection{Medical Relation Extraction Tasks}

Various RE tasks have been undertaken in the medical domain for different purposes.
For instance, 
\citet{parikh-etal-2019-browsing} aimed at improving access to medical information and 
\citet{wolf-etal-2019-term} tackles entity extracting from trustworthy medical literature for question-answering assistants.
Dialogue-based entity extraction tasks designed to assist in electronic medical record (EMR) entry \cite{jeblee-etal-2019-extracting,xia-etal-2022-speaker} have all been explored.
More complex tasks include extracting predefined medical entities and their conditions \cite{gao-etal-2023-dialogue-medical, cheng-etal-2022-jamie, yang-etal-2023-learning-leverage} and extracting findings and characteristics from radiology reports \cite{park-etal-2024-novel}.

While recent LLMs have demonstrated the ability to perform RE as a generation task in general domains \cite{wadhwa-etal-2023-revisiting,wan2023gptre}, 
there are few studies applying LLMs to medical RE, focusing only on temporal relations between diseases \cite{kougia-etal-2024-analysing} or drug-related RE \cite{bhattarai-etal-2024-document}.
While these studies focus on the conditions of medical entities, CTE is unique in its focus on the causal relationships between higher-level diseases.

For collecting data on causal relationships between diseases and findings, \cite{khetan-etal-2022-mimicause} proposed a dataset with annotation specifications covering four types of causal relationships between diseases.
Compared to CTE annotation specification, it differs because CTE constructs a tree structure and extract primary diseases as root.

\section{Conclusion}
% We propose a novel causal tree extraction task that organizes the logical structure of diagnoses and condenses text length to enhance the comprehensibility of medical case reports with the corresponding collected dataset named J-Casemap. 
% We proposed an LLM-based generation model with continual pretraining on medical corpora and fine-tuning on the J-Casemap dataset to achieve accurate summarization.
% Human evaluation suggests that our generation model can achieve a high score of 82.7 points on the 0-100 scale.
% In the subsequent investigation, we substantially improved the correlation between automatic and human evaluation through a heuristic weighting method, allowing the automatic scores to reflect human experts' preferences better.

% Our automated causal trees, with slight human post-editing, is expected to accelerate J-Casemap data accumulation significantly. 
% Imagine an exciting blueprint of a doctor diagnosing a new case by quickly searching large-scale historical J-Casemap data for similar medical findings and tracing their causes back through the tree structure, enabling rapid and accurate diagnostic decisions.

% \textcolor{blue}{
% 私たちは専門家のような文章理解を要する新しいタスク、因果木要約を提案し、症例報告とその因果木要約を含むJ-Casemapのデータセットを構築した。
% また、J-Casemapを用いたLLMのSFTを行うことで因果木要約タスクに取り組み、複数の評価手法において既存手法を上回るスコアを獲得した。
% さらに、ヒューリスティックな重みづけによって自動評価手法を改善し、より人間の専門家の思考を反映した自動評価スコアの算出を可能にした。
We proposed a novel task, causal tree extraction (CTE), which requires expert-like text comprehension, and we constructed the J-Casemap dataset containing case reports and their causal trees.
We tackled the CTE task by fine-tuning LLMs and achieved higher scores than existing methods across both automatic and human evaluations.
Furthermore, we improved the automatic evaluation through heuristic weighting, which reflects clinicians' preferences in automatic evaluation scores.
% }

% \textcolor{blue}{
% 症例報告の因果木は、医学タスクの訓練にも有効であることが示された。
% 因果木による俯瞰的かつ効率的な構造化が提供する、高度な因果推論についての知見は、医学以外の分野にも応用できる可能性を秘めている。
The causal tree of case reports is useful not only for clinicians but also for LLMs to train along with other medical tasks, such as question answering tasks.
The insights into advanced causal reasoning have the potential to be applied in domains beyond medicine.

\section{Limitations}
% ##### Limitation
% In this study, the effectiveness of the J-Casemap dataset in assisting actual diagnoses by clinicians was only partially verified.
% The J-Casemap dataset is currently limited to members of the Japanese Society of Internal Medicine, making it difficult to make the data publicly available.
% The hyperparameters for the experiments were manually determined, suggesting room for improvement.
Hallucination problems were seen in the LLMs' outputs, but we have not discussed the solutions in this paper.
In future work, more advanced approaches like Retrieval-augmented generation or entity linking between the causal tree and the case report text are probably needed to find the supporting evidence towards more reliable generation.

% 内科以外
Besides, all of the case report data in this experiment are from internal medicine, which potentially limits the scope of this study.
We are ambitious in envisioning the future where the J-Casemap data is expanded beyond internal medicine to other departments, ultimately establishing a unified standard across different medical fields. 

% ##### Future Works
% Future research plans include conducting SFT with larger models to improve the quality of automatic generated causal trees.
The last limitation lies in the automatic evaluation of CTE. Even though we already improved automatic metrics, developing more comprehensive and accurate automatic metrics that more closely resemble manual evaluation is necessary.
% Moreover, we plan to verify the effectiveness of using J-Casemap as training data for LLMs.

\section{Ethical Statement}
The copyright of the J-Casemap dataset belongs to the Japanese Society of Internal Medicine, making it difficult to make the data publicly available due to privacy and security concerns.
we will release the final version of the annotation schema and 100 causal tree samples based on public case reports without ethical concerns from the Japan national medical license examination.

\section*{Acknowledgments}

\bibliography{main}

\appendix

\section{Searching optimal triplet weights}\label{sec:appendix-triplet-weight} 
We design two weighting methods for the triplet evaluation as follows:

\noindent Weighting method 1: $$ W = \frac{1}{1+Cd}x_{relation} $$
$ x_{relation} $ is $ 1 $ when the relation type is $ parent\_of, $ and $\frac{1}{2}$ if not.

\noindent Weighting method 2: $$ W = \frac{1}{C^{d}}x_{relation} $$
$ x_{relation} $ is $ 1 $ when the relation type is $ parent\_of, $ and $\frac{1}{C}$ if not.
$C$ is a constant hyper-parameter that can be tuned. $d$ is the triplet depth. 

\begin{figure}[t]
\centering
\includegraphics[width =1.0\linewidth]{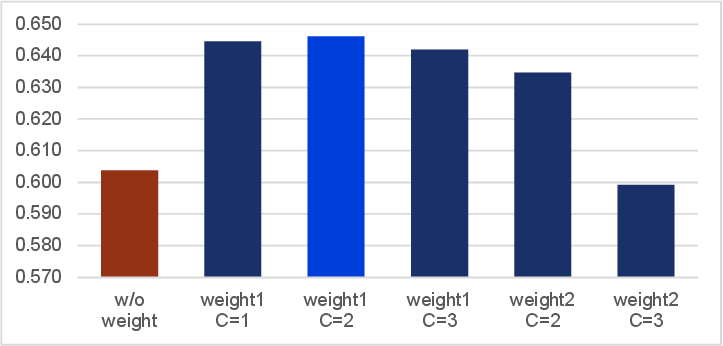}
\caption{Correlation between manual scores and automatic scores. 600 causal trees generated by the RE model and generation model for 300 case reports were automatically evaluated, and correlation coefficients with human scores were calculated.}\label{fig:correlation}
\end{figure}

We further calculate the correlation coefficients of weighting factors in automatic evaluations, shown in Figure \ref{fig:correlation}.
It was noticed that the weighting of prioritized entities in lower layers showed a higher correlation with manual evaluations.
However, when extreme weighting was applied, the correlation with manual evaluations decreased. 
The Appendix \ref{sec:appendix-eval} provides a more detailed analysis.

After we assign heuristic weights to the automatic evaluation, the performances become closer to the human clinicians, as shown in Table \ref{tab:manual-evaluation}. 
Currently, automatic evaluation is still unable to match human doctors' precision in judging salient information and ideally identifying entities.
We consider this an open issue for future research.
The weighting pattern $1$ ($C=2$), which shows the highest correlation coefficients to human scores, is used in all the following experiments.

\section{Comparison of Model Prediction Trends}\label{appendix:comparison-methods}
This section provides a more detailed analysis and comparison of the RE model and the best generation model, Med-swallow-13b.
Statistics on the number of generated triplets and root nodes are shown in Table \ref{tab:triplets}.
Compared to RE models, generation models extracted more triplets and had fewer omissions in information extraction.
Additionally, the RE model predicts far more root nodes than Gold, while the Generation model predicts about the same number of roots as Gold.
This indicates that the generation model was able to designate a few critical entities as root elements and link other entities comprehensively downstream.
On the other hand, the RE model enumerated extracted entities that did not have identified relationships as root elements.
These aligns with the experimental results that showed a significant difference in precision and a smaller difference in recall.

\begin{table}[t]
\centering
\begin{tabular}{crr}
\hline
           & \multicolumn{1}{c}{Triplets} & \multicolumn{1}{c}{Root node} \\ \hline
Gold       & 14,049                       & 545                           \\
RE model         & 13,343                       & 1,584                         \\
Generation model & 14,453                       & 550                           \\ \hline
\end{tabular}
\caption{The statistics on the number of triplets and root nodes. Med-swallow-13b is used as generation model.}\label{tab:triplets}
\end{table}

\begin{figure}[t]
\includegraphics[width =1.0\linewidth]{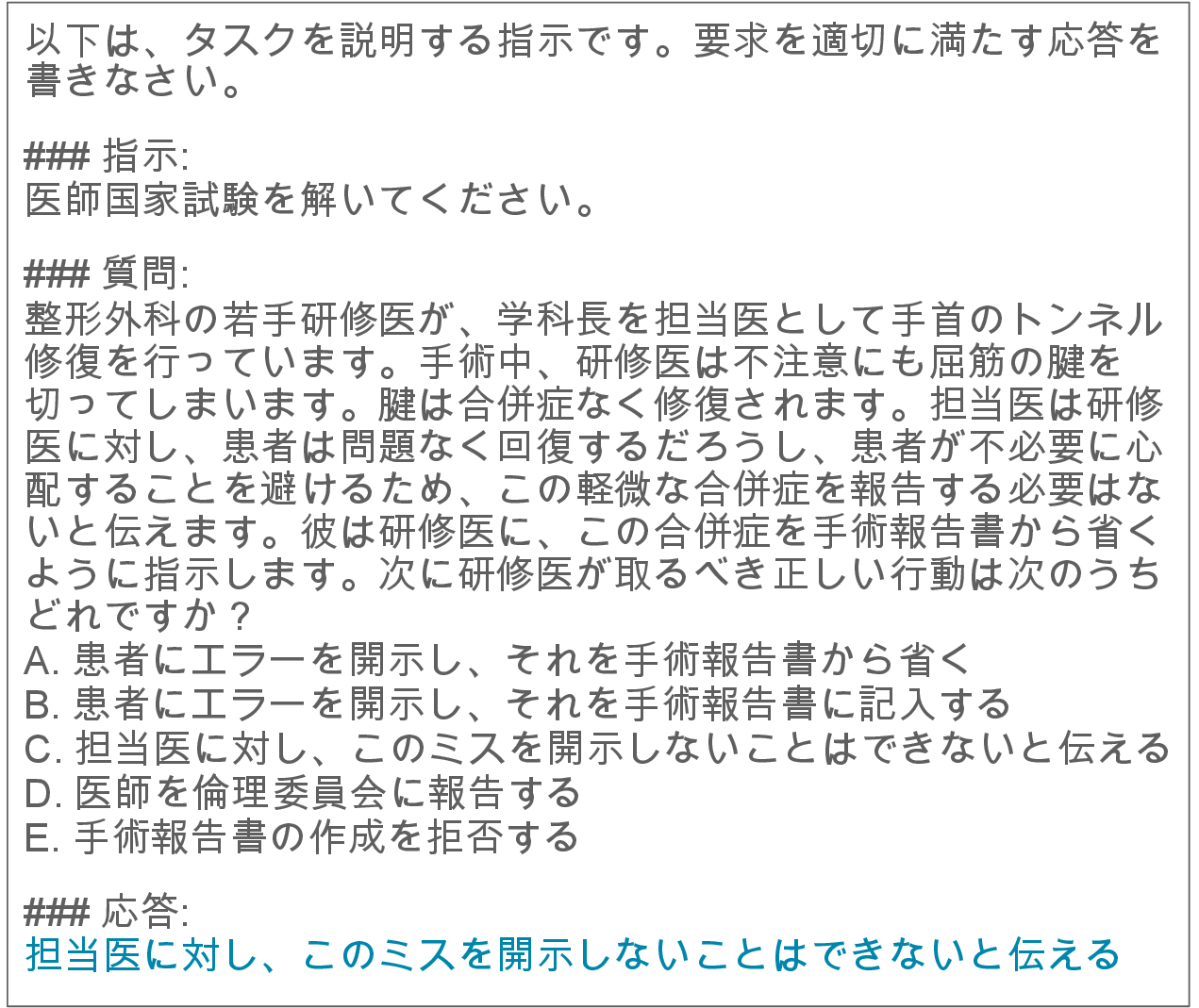}
\caption{MedQA prompt for SFT. The blue parts were used for loss calculation.}\label{fig:MedQA prompt}
\end{figure}

\begin{table}[t]
\centering
\begin{tabular}{lc}
\hline
           \multicolumn{1}{l}{Hyper-parameters}& \multicolumn{1}{c}{Value} \\ \hline
Constant learning rate       & $3.00e^{-6}$                                             \\
Warm-up schedule         & Linear                                               \\
Warm-up ratio & $0.03$                                                 \\ 
Weight decay & $0.1$                                                 \\
Data type & bf$16$                                               \\
Global batch size & $32$                                                 \\
\hline
\end{tabular}
\caption{Hyper-parameters of pretraining}\label{tab:pretrain}
\end{table}

\section{Prompt Template for Different LLMs}
\label{appendix:LLMprompt}
% The template for the prompts used in the experiment is shown in the figure \ref{}.
Due to differences in model compatibility, two types of inference templates were used according to the model.
The inference templates follow the examples provided on the model card for each model.
Additionally, a beginning-of-sequence (BOS) token was added at the start of the prompt, and an end-of-sequence (EOS) token was added at the end of the LLM-generated outputs during training and testing.

\section{MedQA prompt}\label{appendix:MedQA prompt}
We present the prompt for MedQA SFT in Figure \ref{fig:MedQA prompt}.

\section{Hyperparameters}
\label{appendix:hyper}

\begin{table*}[ht!]
\centering
\begin{tabular}{ccc}
\hline
\begin{tabular}{r}{}model\end{tabular} & \begin{tabular}[c]{@{}c@{}}LLM-jp-13b-v1\\Med-llm-jp-13b-v1\end{tabular} & \begin{tabular}[c]{@{}c@{}}Swallow-13b \\ Med-swallow-13b\end{tabular}                      \\ \hline
batch-size                                                     & 64                                                                                                                 & 64                                                                                                                           \\
max\_seq                                                      & 2048                                                                                                               & 4096                                                                                                                         \\
learning rate                                                 & 1.00E-04                                                                                                           & 1.00E-04                                                                                                                     \\
warmup ratio                                                  & 0.1                                                                                                                & 0.1                                                                                                                          \\
LoRA target modules & c\_attn, c\_proj, c\_fc                                                 & \begin{tabular}[c]{@{}c@{}}q\_proj, k\_proj, v\_proj, o\_proj,\\ gate\_proj, up\_proj, down\_proj, lm\_head\end{tabular} \\
LoRA alpha                                                    & 32                                                                                                                 & 32                                                                                                                           \\
LoRA r                                                        & 8                                                                                                                  & 8                                                                                                                            \\
LoRA dropout                                                  & 0.05                                                                                                               & 0.05                                                                                                                         \\ \hline
\end{tabular}
\caption{Hyper-parameters of fine-tuning.}\label{tab:hyper-sft}
\end{table*}

We present the detailed hyper-parameters of the pretraining in Table~\ref{tab:pretrain} and the fine-tuning stage in Table~\ref{tab:hyper-sft}.

\section{Case Study of Evaluation Comparison} \label{sec:appendix-eval}

\begin{figure*}[]
\centering
\includegraphics[width=1.0\linewidth]{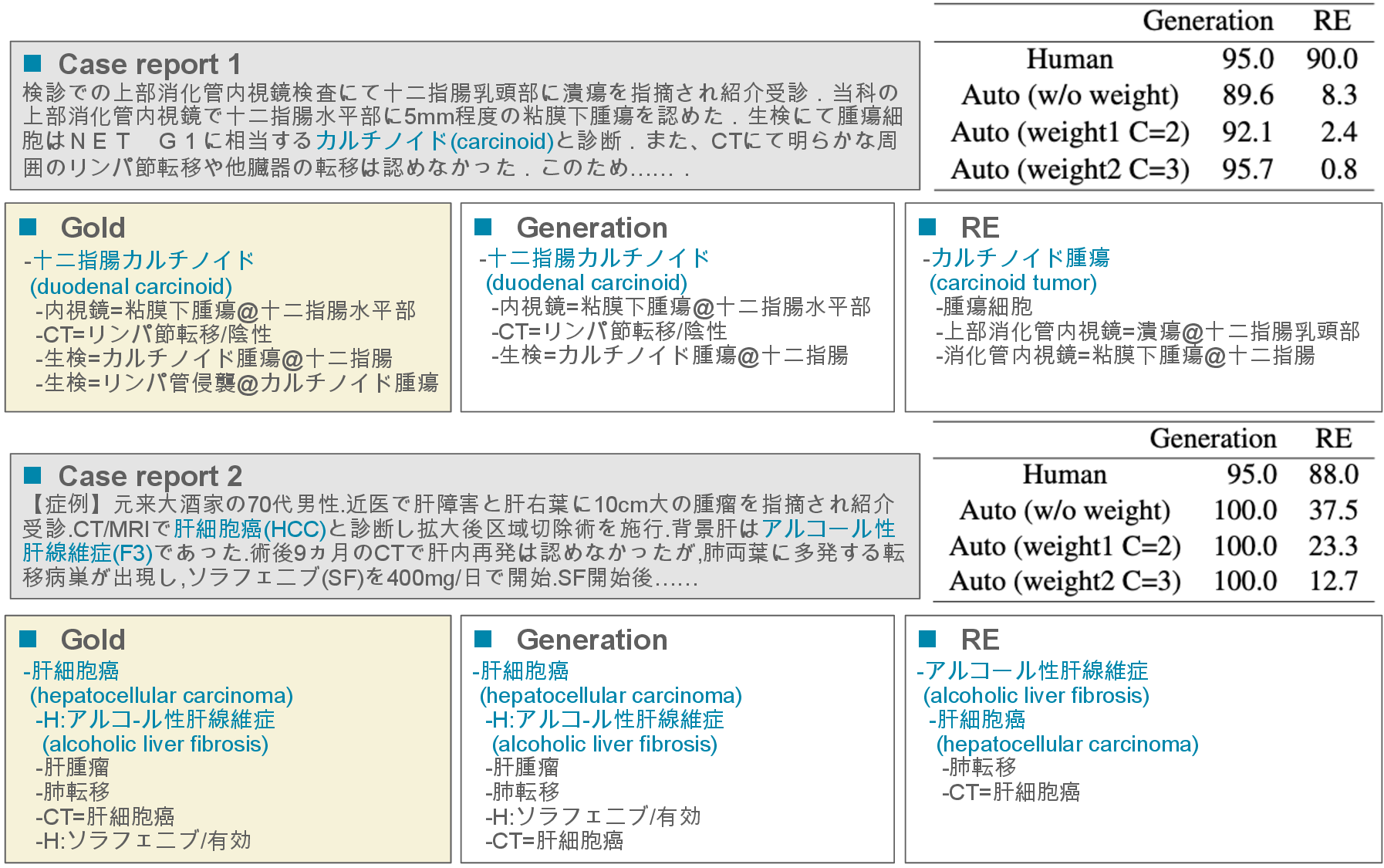}
\caption{Case study of evaluation.}\label{fig:case-study-evaluation}
\end{figure*}

Examples of a case study that focuses on automatic evaluation are shown in Figure \ref{fig:case-study-evaluation}.
In both of the examples, the generated summary of the RE model got good scores in the human evaluation, but the automatic evaluation score is very low.

The reason for the evaluation failure of the case 1 is that the influence of matching errors for entities in lower layers becomes too significant, leading to a lower correlation with the manual evaluation.
While manual evaluations can perfectly match entities, automatic evaluations may fail to do so.

The reason for the evaluation failure of the case 2 is the ambiguity of the causal relationship.
It is occasionally difficult to determine which is the cause and which is the result of the causal relationship between diseases, especially when multiple diseases are combined.

Even with the most correlated weighting, the correlation coefficient remained around 0.6, indicating a substantial gap between manual and automatic evaluation scores.
To perform automatic evaluation more similar to human evaluation, a more flexible evaluation method than evaluation by triplet comparison is required.

\end{document}